\documentclass{article}

\usepackage[accepted]{aistats2024}
\usepackage[round]{natbib}
 \renewcommand{\bibname}{References}
\renewcommand{\bibsection}{\subsubsection*{\bibname}}

\usepackage{hyperref}
\usepackage{url}

\usepackage{microtype}
\usepackage{graphicx}
\usepackage{booktabs} 
\usepackage{subcaption}
\usepackage{amsmath}
\usepackage{amssymb}
\usepackage{mathtools}
\usepackage{amsthm}

\usepackage{makecell}
\usepackage[capitalize,noabbrev]{cleveref}

\theoremstyle{plain}
\newtheorem{theorem}{Theorem}[section]

\theoremstyle{definition}
\newtheorem{definition}[theorem]{Definition}

\theoremstyle{remark}

\def\R{{\mathbb{R}}}

\usepackage{hyperref}
\usepackage{url}
\usepackage{algorithm}

\usepackage{algpseudocode}
\usepackage{booktabs}
\usepackage{adjustbox}
\usepackage{ amssymb }
\usepackage{multirow}

\usepackage{amsmath,amsfonts,bm}

\def\eqref#1{equation~\ref{#1}}

\def\1{\bm{1}}

\def\rl{{\textnormal{l}}}

\def\ru{{\textnormal{u}}}

\def\rx{{\textnormal{x}}}

\def\rvu{{\mathbf{i}}}

\def\rvl{{\mathbf{l}}}

\def\rvu{{\mathbf{u}}}

\def\rvx{{\mathbf{x}}}

\DeclareMathAlphabet{\mathsfit}{\encodingdefault}{\sfdefault}{m}{sl}
\SetMathAlphabet{\mathsfit}{bold}{\encodingdefault}{\sfdefault}{bx}{n}

\def\sA{{\mathbb{A}}}

\def\sI{{\mathbb{I}}}

\def\sS{{\mathbb{S}}}

\def\sX{{\mathbb{X}}}

\DeclareMathOperator*{\argmax}{arg\,max}

\usepackage[textsize=tiny]{todonotes}

\begin{document}

%

%

\twocolumn
[\aistatstitle{Guarantee Regions for Local Explanations}

\aistatsauthor{Marton Havasi \And Sonali Parbhoo \And Finale Doshi-Velez }

\aistatsaddress{Harvard University \And  Imperial College London \And Harvard University } ]

\begin{abstract}
Interpretability methods that utilise local surrogate models (e.g. LIME) are very good at describing the behaviour of the predictive model at a point of interest, but they are not guaranteed to extrapolate to the local region surrounding the point. However, overfitting to the local curvature of the predictive model and malicious tampering can significantly limit extrapolation. We propose an anchor-based algorithm for identifying regions in which local explanations are guaranteed to be correct by explicitly describing those intervals along which the input features can be trusted. Our method produces an interpretable feature-aligned box where the prediction of the local surrogate model is guaranteed to match the predictive model. We demonstrate that our algorithm can be used to find explanations with larger guarantee regions that better cover the data manifold compared to existing baselines. We also show how our method can identify misleading local explanations with significantly poorer guarantee regions.
  
\end{abstract}

\section{Introduction}
Explanations of machine learning models are becoming increasingly important across industries in order to aid human understanding, increase transparency, improve trust, and to better comply with regulations. Currently, local surrogate models are widely employed for this task. Informally, surrogate models are simpler models with fewer parameters (such as logistic regression or decision trees) used for explaining the behaviour of a more complex model around a point of interest \citep{ribeiro2016should}. However, local surrogates can often be misleading since it may be unclear how far one can extrapolate the prediction of the surrogate model while accurately matching the predictions of the complex model \citep{slack2020fooling}. This issue motivates the need for interpretable methods that can bound the faithful region (shown in Figure 1 (Left)), where the prediction of a surrogate is within a specified error tolerance of the prediction of a complex model, to prevent incorrectly extrapolating local explanations. For instance, we require that at least $\rho=99\%$ of the volume of the bounded region is faithful with confidence at least $1-\delta =99\%$. 

A natural choice for bounding the faithful region of the input space where an explanation applies is using an anchor box (also referred to as a box explanation or an anchor explanation) \citep{ribeiro2018anchors}. An anchor box specifies the interval along each input feature in which the input point must fall as shown in Figure 1 (Right). Its coverage is clear: given a point $\rvx$, one simply needs to compare each of the $D$ features against the lower and upper bounds ($\rvl$ and $\rvu$) of the respective anchor box. $\rvx$ is within the anchor box when $\rl_d \leq \rx_d \leq \ru_d$ for $d=1\dots D$. Yet computing the largest anchor boxes in a continuous domain is an intractable task for which no solution currently exists. 

\begin{figure}[t]
\centering
\includegraphics[scale=0.3]{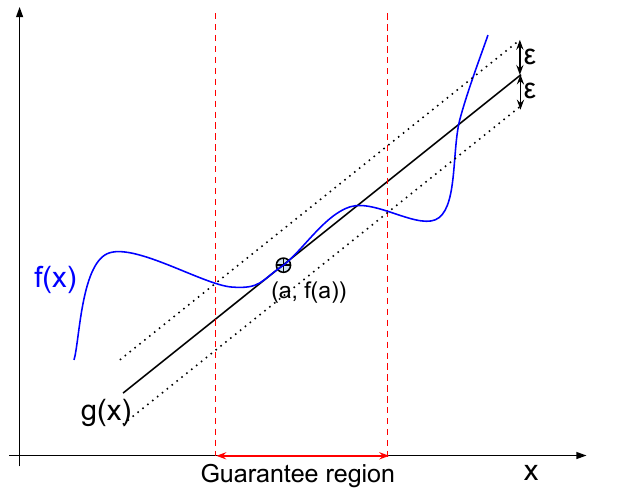} \label{fig:1d_plot}
\includegraphics[scale=0.3]{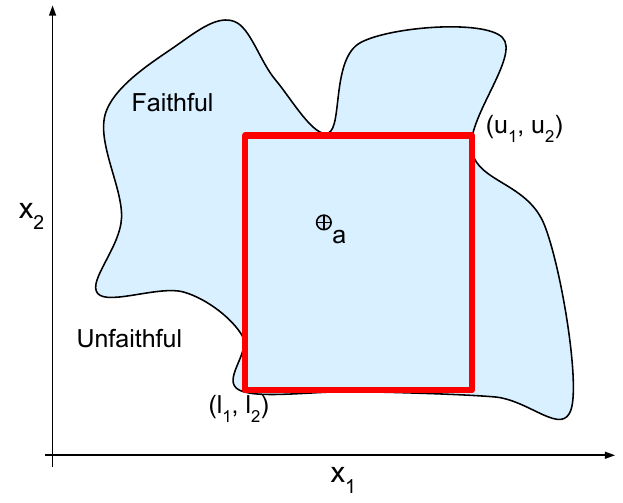}\label{fig:anchor_plot}
\caption{(Left) The guarantee region captures the area around the anchor point $a$ in which the prediction of the complex model $f$ differs from the surrogate model $g$ by no more than $\epsilon$. (Right) Anchor box of the faithful region in $D=2$ dimensions (shown in red). The anchor box captures an axis-aligned box (defined by lower and upper bounds $l=(l_1, l_2)$ and $u=(u_1, u_2)$) where the surrogate model is guaranteed to be faithful to the complex model. }
\end{figure}
In this work, we prove that approximating the largest box requires at least $O({D \choose \lfloor \frac{D}{2} \rfloor})$ model evaluations in the worst case -- making it infeasible even for moderate ($D=10-30$) dimensional data. 

Yet many applications of machine learning use data with moderate dimensionality for making inferences. E.g. In healthcare, treatements are typically prescribed off a low-dimensional subset of relevant measurements with relatively few degrees of freedom. We propose a new divide-and-conquer algorithm \emph{with statistical guarantees} that breaks the task of finding the maximal anchor box into smaller subproblems. The anchor box is initially based on a subset of the $D$ features, and then incrementally built up to the full anchor box that covers all features. Our experiments show i) if a local explanation claims a feature is not important for a complex model, our anchor box can detect this by showing that the local explanation does not extrapolate along the said feature; ii) our method finds significantly larger guarantee regions that better cover the data manifold than baseline approaches. 


\section{Related Works}
\textbf{Local explanations.} Local approaches to interpretability provide an explanation for a specific input.  These methods trade off the ease of an explanation with its coverage: Though an explanation may be simple for complex tasks, it may only explain the behaviour of a model in a localized region of the input space. \cite{ribeiro2016should} showed that the weights of a sparse linear model can be used to explain the decisions of a black box model in a small area near a fixed data point. Linear functions can capture the relative importance of features easily but may not generalise to unseen instances. Alternatively, \cite{singh2016programs} and \cite{koh2017understanding} use simple programs or influence functions respectively to explain the decisions of a black box model in a localized region. Other approaches have used input gradients (which can be thought of as infinitesimal perturbations) to characterize local logic \cite{van2008visualizing,selvaraju2016grad}. In this paper, we focus on bounding the region in which these explanations are faithful.

\textbf{Methods for validating explanations.} Among various studies aimed at validating explanations, \cite{Ignatiev2019OnVR} demonstrate that most existing techniques for explainability are heuristic in nature, and that counterexamples of these heuristics can expose some of the defects of the methods. Alternatively, \cite{ijcai2018p708,10.1609/aaai.v33i01.33011511} propose a method to computing provably correct explanations in Bayesian network classifiers. Unlike these works, we focus on explicitly identifying near-optimal anchor boxes with the largest coverage while fulfilling the required statistical guarantees. 

\textbf{Anchor explanations.}  Local, model-agnostic explanations known as \emph{anchors} \cite{ribeiro2018anchors} provide explanations that sufficiently ``anchor” a prediction locally, such that changes to the feature values of the instance do not affect the prediction. They are easy to understand, while providing high precision and clear coverage: they only apply if all the conditions of a rule are satisfied (see \cite{ribeiro2018anchors}). However, prior work on anchor boxes only work with categorical features. Our work further develops anchor boxes, shows how they can be adapted to continuous settings and proposes an algorithm for finding those anchor boxes with the largest coverage.

\textbf{Max-box methods.} A problem closely related to anchor boxes is the max-box problem. Given a set of negative and positive points, one must find an axis-aligned box that contains as many positive points as possible while avoiding any negative points. Many variants of this problem have been researched for computer graphics/computational geometry, however, the results rarely extend to $D>2$ dimensions \citep{liu2003planar,atallah1986note,barbay2014maximum}. \citet{eckstein2002maximum} showcase an algorithm for $D>2$ dimensions that frames the question as a search problem. Our paper builds on this work and proposes an algorithm for the problem variant where there are no positive and negative points available, we merely have a function that we can evaluate to see if a point is positive (i.e. faithful) or negative (i.e. not faithful). \citet{dandl2023interpretable} propose a method for obtaining interpretable regional descriptors (IRDs) that are model agnostic. IRDs can provide hyperboxes which describe how an observation’s feature values can be changed without affecting its prediction. The authors justify a prediction by providing a set of semi-factual statements in the form of "even-if" arguments and they show which features affect a prediction. Unlike Max box methods - a class of IRDs - we demonstrate that our approach can be applied to higher dimensions. \citet{sharma2021maire} propose a method (MAIRE) to find the largest coverage axis aligned hypercuboid such that a high number of instances within the hypercuboid have the same class label as the instance being explained (high precision). The authors focus on approximating coverage and precision measures and maximise these using gradient-based techniques. Though similar in its use of hypercubes to explain the label of an instance, unlike \citet{sharma2021maire}, we develop a divide-and-conquer algorithm for finding a maximum anchor box by successively capturing only the positive points and pruning out the negative ones. In our experiments, we also provide evidence and analyses on how much of a local cluster can be captured using our method.  Unlike all these works, to the best of our knowledge we are also the first to show that the problem cannot be solved tractably even for moderately sized problems.

\section{Anchor Boxes with Statistical Guarantees}
\textbf{Notation and Setting.} An anchor box $\sA$ for continuous data of dimension $D$ is defined by a vector of lower bounds $l$ and a vector of upper bounds $u$, that is, $\sA \triangleq (l, u)$. A point $x \in \R^D$ is contained in anchor box $sA$ if it falls within the bounds along all $D$ dimensions: $
x \in \sA \; \text{ if } l_d \leq x_d \leq u_d \; \forall d \in D \,.$
We are focused on situations in which the explanation is computable; that is, the explanation of some original function $f(x)$ is a simpler surrogate function $g(x)$.  For any input $x$, let $e(x) \rightarrow \{1, 0\}$ indicate whether the explanation $g(x)$ is sufficiently faithful to the original function $f(x)$:\footnote{In the classification setting, one could define similarity akin to \cite{ribeiro2018anchors}, where
$e(x)$ indicates when $g(x)$ is adequate to predict a specific class $f(x)$
}
\begin{align}
e(x)=
\begin{cases}
1 \quad \text{if} \, |f(x) - g(x)| < \epsilon  \\
0 \quad \text{otherwise} \,.
\end{cases}
\end{align}
where $\epsilon$ is a user-defined quantity regarding how close we require the model and the explanation to be to consider the explanation faithful enough.  Finally, an anchor box $\sA$ has \emph{purity} 
$\rho$, if $\mathbb{P}_{U(\sA)}(e(x)=1) = \rho$ where $U(\sA)$ denotes the uniform distribution over the points in $\sA$.

Conceptually, the explanation $g(x)$ has some region around $x$ for which it explains the original function $f(x)$ faithfully.  A good anchor box covers as much of this region as possible---it does not leave out situations in which $g(x)$ matches $f(x)$.  Formally, we consider an anchor box better than another when it covers a larger volume at the same purity level, where the volume of a box $\sA$ is given by $|\sA|=\prod_{d=1}^D u_d-l_d$. An advantage to this notion of quality is that it is invariant to scaling and shifting: a larger anchor box remains larger when features are scaled and shifted, so there is no concern if some features operate on a drastically different scale than others. Note that none of our definitions require knowledge of the data-generating distribution.  However, in practice, we do require knowledge of maximum and minimum values for each feature, denoted by the vectors $\mathsf{ub}$ and $\mathsf{lb}$.  Requiring the anchor box $\sA$ to lie inside these maximum and minimum values prevents cheating by setting $l_d=-\infty$ and $u_d=\infty$ for an irrelevant feature $d$ to get infinite volume. In our experiments, we set the minimum and maximum values for each feature at the minimum and maximum values in the data.

\subsection{The Computational Challenge: Identifying Maximal Anchor Boxes Grows Exponentially in Dimension}
We want to find the maximum anchor box, that is, output the full (axis-aligned) region for which the explanation $g(x)$ matches the function $f(x)$.  Here, we prove a result in Theorem \ref{theorem:exp} that states that the number of computations required to reliably approximate that maximum anchor box grows exponentially in dimension, making it impossible to identify maximal anchor boxes in higher dimensions. 

\begin{definition} \textbf{Maximum Anchor Box.}  Given some anchor point $x^*$, an indicator of similarity $e(x)$, and a desired purity threshold $\rho$, the maximum anchor box $\sA_\text{max}(x^*, e(x), \rho)$ is the largest box containing $x^*$ that has purity at least $\rho$ with respect the similarity indicator $e(x)$:
\begin{align}
\sA_\text{max}(x^*, e(x), \rho) = \argmax_{\sA=(l, u)} \prod_d^D u_d - l_d \\\text{s.t.}\nonumber;  \mathbb{P}_{U(\sA)}(e(x)=1) \geq \rho \;\text{and}\; x^* \in \sA
\end{align}
\end{definition}
Next we show the number of evaluations needed to approximate the maximum anchor box grows exponentially with the number of dimensions, making it intractable to compute in practice. Then we introduce an algorithm to find a reasonable approximation of the maximum anchor box. The impossibility result is a consequence of the rapidly increasing number of possible anchor boxes with dimension $D$. As the dimension $D$ grows, there may be very large number of anchor boxes $\sA$ that overlap very little, thus requiring us to test the purity of each independently. Formally:

\begin{theorem}
\label{theorem:exp}
If there exists a true function $e_0$ for the data and a set of possible faithfulness functions $\mathcal{E}=\{e_k: \R^D\rightarrow \{0, 1\}\}_{k=1}^K$ with $K={D \choose \lfloor \frac{D}{2} \rfloor}$ that agree on a collection of input points, but have different bounding boxes and  $\frac{|\sA_\text{max}(x^*, e_0, \rho=1)|}{|\sA_\text{max}(x^*, e_k, \rho=1)|} = r$, where $0<r<1$ is the ratio of volumes of maximal bounding boxes for different $e_k$, then for $L$ evaluations, $x^{(1)} \dots x^{(L)}$, 
\begin{align}
    |\{e_k \in \mathcal{E}|\forall x^{(l)}. e_0(x^{(l)}) = e_k(x^{(l)}) \}| \geq K-L
\end{align} 
Proof: See Appendix \ref{app:proof} for details.
\end{theorem}


Theorem \ref{theorem:exp} states that if there is a set of functions $\mathcal{E}$ that is equal to $e_0$ for a large portion of the input space then if we only have $L$ evaluations, one can distinguish at most $L$ functions from $e_0$. If the target function $e_k$ is chosen randomly from $\mathcal{E}$ $(e_k\sim U(\mathcal{E}))$ after $L$ evaluations, the probability of identifying the correct $e_k$ is at most $\frac{L}{K}$ (without identifying $e_k$, one has to fall back to $\sA_{\text{max}}(x^*, e_0, \rho=1)$ which is worse than $\sA_{\text{max}}(x^*, e_k, \rho=1)$ by a factor of $r$). As a result, we need $\mathbb{E}[L]=\frac{K}{2}$ function evaluations in expectation to identify $e_k$. Since $K\geq{D \choose \lfloor \frac{D}{2} \rfloor}$ we are able conclude that any algorithm that claims to find the maximum anchor box up-to a factor $r$ needs on average at least $O({D \choose \lfloor \frac{D}{2} \rfloor})$ function evaluations in the worst-case problem setting. Specifically, we show that one needs at least $O({D \choose \lfloor \frac{D}{2} \rfloor})$ function evaluations of the error function $e$ to identify the maximum anchor box $\mathbb{A}_{\text{max}}$ to approximate the largest possible anchor box to a factor $0 < r < 1$ with $\rho=1$. When $\rho \neq 1$, a similar issue if not worse arises.

\begin{algorithm}[t]
\caption{$\mathsf{FindAnchor}(\sI, l, u)$}\label{alg:recursion}
\begin{algorithmic}
\If{$|\sI|=1$}
    \State {\bfseries return} $\mathsf{SolveRestricted}(\sI, l, u)$
\Else
    \State $\sI_1, \sI_2 \gets \mathsf{BalancedRandomSplit}(\sI)$ 
    \State $l_1, u_1 \gets \mathsf{FindAnchor}(\sI_1, l, u)$
    \State $l_2, u_2 \gets \mathsf{FindAnchor}(\sI_2, l, u)$
    \State $l, u \gets \min(l_1, l_2), \min(u_1, u_2)$
    \For{$i=1 \dots \min(|\sI_1|, |\sI_2|)$}
        \State $l, u \gets \mathsf{SolveRestricted}(\sI_1 \cup \sI_2[:i], l, u)$
        \State $l, u \gets \mathsf{SolveRestricted}(\sI_2 \cup \sI_1[:i], l, u)$
    \EndFor
    \State {\bfseries return} $l, u$
\EndIf
\end{algorithmic}
\end{algorithm}

\begin{algorithm}[t]
\caption{$\mathsf{SolveRestricted}(\sI, l, u)$}\label{alg:solution}
\begin{algorithmic}
\State $\sX^+, \sX^-  \gets \{\}, \{\}$
\While{$|\sX^+| < N$}
    \State $x\sim U(l, u)$
    \State $x_i \gets a_i \;\text{ for }\; i\notin \sI$
    \If {$b(x)=1$}
        \State $\sX^+ \gets \sX^+ \cup \{x\}$
    \Else
        \State $\sX^- \gets \sX^- \cup \{x\}$
    \EndIf
\EndWhile
\For{$i=1\dots \infty$}
\State $l', u' = \mathsf{FindMB}(\sX^+, \sX^-, \sI, l, u)$
\State $\delta_i \gets \frac{\delta}{ i \log^2(i+1) \sum_{j=1}^\infty\frac{1}{j\log^2(j+1)}}$
\State $M \gets \lceil \frac{\log(\delta_i)}{\log(\rho)}\rceil$
\State $\triangleright$ Comment: $M$ positive random samples are needed for a statistical test with confidence $1-\delta_i$
\State $x^{(1)}\dots x^{(M)} \sim U(l', u')$
\If{$\exists \; e(x^{(m)})=0$}
\State $\sX^- \gets \sX^- \cup \{x^{(m)}|e(x^{(m)})=0\}$
\Else
\State {\bfseries return} $l', u'$
\EndIf
\EndFor
\end{algorithmic}
\end{algorithm}


\subsection{A Divide-and-Conquer Algorithm for Finding Large Anchor Boxes}
Computing the maximum anchor box is intractable in high dimensions. Here, we present the key contribution of our work: an approximate algorithm that finds a large anchor box in which local explanations are guaranteed to match the predictive model of interest. The challenge lies in the dimensionality of the problem. At dimensions $D=10$-$30$, the largest anchor box only covers a tiny portion of the input space. Moreover, there are many possible large anchor boxes that overlap very little. 

\textbf{Overview}
We develop a divide-and-conquer strategy to find large anchor boxes.  First, we find one-dimensional anchor boxes for each input feature $d$---that is, only $x_d$ may vary, and the other dimensions much match $x^*$---that have our desired purity $\rho$.  Next, we use these one-dimensional anchor boxes to determine anchor boxes where two features may vary simultaneously. Knowing the one-dimensional boxes enables us to restrict the search space of two-dimensional boxes to only those that lie within the bounds of the one-dimensional boxes associated with those two input features. This strategy is repeated in succession to build an anchor box that eventually covers all dimensions, whilst avoiding a prohibitively large search space.

\textbf{Base Case and Recursion}
Let us define an anchor box $\sA_\sI$ as an anchor box that varies across some subset of dimensions $\sI \subseteq \{1\dots D\}$.  That is, the anchor box contains points $x$ such that $l_d \leq x_d \leq u_d$ for dimensions $d\in \sI$ and $x_d = x^*_d$ for 
$d \notin \sI$. Our divide-and-conquer strategy forms subsets by recursively halving the feature set $\sI$ (the halves are allocated randomly, and when the number of features is odd, one half has an extra feature). The base case contains a single feature, that is, $\sI = \{d\}$.

Now let us consider the recursion: we have large anchor boxes for some subsets of dimensions $\sI_1$ and $\sI_2$.  Now, we want to find a large anchor box for $\sI=\sI_1 \cup \sI_2$.  The first thing we do is restrict the overall maximum and minimum values of the dimensions: $\mathsf{lb_d}, \mathsf{ub_d}\gets l_d, u_d$ for $d\in \sI$ based on the solutions $\sA_{\sI_1}$ and $\sA_{\sI_2}$.

When the sets $\sI_1$ and $\sI_2$ consist of many dimensions, it is important that process of merging solutions reduces the search space gradually---trying to immediately identify the anchor box for the space $\sI=\sI_1 \cup \sI_2$ would result in a disproportionately large search space compared to the size of the anchor box.  Thus, below we will merge the dimensions one by one successively adding the elements of $\sI_2$ to $\sI_1$ and subsequently $\sI_1$ to $\sI_2$ (i.e. we first solve $\sI_1\cup\sI_2[:1]$, then $\sI_2\cup\sI_1[:1], \; \sI_1\cup\sI_2[:2], \dots, \sI_1 \cup \sI_2$ where $\sS[:i]$ denotes the first $i$ elements of $\sS$). Note that though we add elements to $\sI_1$ one at a time, the effort to find the bounding box for $\sI_2$ is still useful since $\sI_2$ considers all the dimensions in $\sI_2$ jointly, making $\sI_2$'s region for any single dimension smaller than if we were to add that dimension on its own.  Pseudocode for this iterative merge process is shown in Algorithm \ref{alg:recursion}. To find the anchor box covering all features, one must call $\mathsf{FindAnchor}(\sI=\{1\dots D\}, l=\mathsf{lb}, u=\mathsf{ub})$, where $(\mathsf{lb}, \mathsf{ub})$ denotes the bounding box.

It remains to define how exactly we add dimensions to an existing anchor box candidate, that is, perform merges such as finding the best anchor box for $\sI_1\cup\sI_2[:i]$.  Our solution will adapt and use as a subroutine an efficient algorithm for the maximum-box problem proposed by 
\cite{eckstein2002maximum} (we refer to their algorithm as $\mathsf{FindMB}$). In the maximum-box problem, there exists a fixed, finite set of negative points $\sX^- \subset \R^D$ and a fixed, finite set of positive points $\sX^+ \subset \R^D$. The task is to find the axis-aligned box containing the most positive points without containing any negative points. While their maximum-box problem and our maximum anchor box problem are not the same, they are closely connected.  If we sampled a very large number of points and designated the positive ones where $e(x)=1$ and the negative ones where $e(x)=0$, then the maximum-box problem (restricted to boxes that contain the anchor point) would coincide with the largest anchor box with purity $\rho=1$.

A detailed description of the $\mathsf{FindMB}$ is found in the original paper \cite{eckstein2002maximum}.  At a high level, $\mathsf{FindMB}$ works by constructing a search tree of possible values for $l$ and $u$. The tree is searched for possible solutions while pruning away branches that are suboptimal. To apply it to our anchor box problem, we make three modifications to $\mathsf{FindMB}$. First, we require the anchor box to contain $x^*$ (we achieve this by simply setting $\bar{l}=\underline{u}=x^*$ using their notation). Second, we do not search the whole tree. We set a stopping criteria that after searching $T$ nodes, the search returns the best box so far. This change is needed to bound the run time when the search tree is very large. Since we add each dimension one by one our computations are bounded in practice. Finally, we want the resulting box to border negative points or the bounding box. This step helps to avoid underestimating the anchor box. We expand the sides of the box until they either meet a negative points or the maximal bounding box. The expansion is done in the order that prioritizes the largest increase in area (in Appendix \ref{app:expansion}, we show that this yields an improvement over a random order of expansion).

The challenge to use $\mathsf{FindMB}$ to find the new anchor box is that in our setting, we do not have a fixed, finite set of positive and negative points. To obtain positive and negative points ($\sX^+$ and $\sX^-$) for $\mathsf{FindMB}$, we first sample points uniformly in the search space and designate them according to $e$ ($e=1\rightarrow$ positive, $e=0\rightarrow$ negative) until we accumulate $N$ positive points. Then, we find the maximum box using $\mathsf{FindMB}$ and apply a statistical test to see it meets the purity requirement. The test at significance level $\delta_i$ requires that $M=\lceil \frac{\log(\delta_i)}{\log(\rho)}\rceil$ random samples from the box all meet $e(x^{(m)})=1$, where below the significance level $\delta_i$ will be chosen in a way to ensure that the final anchor box will have purity $\rho$ with probability at least $1-\delta$ (i.e. we need to compensate for multiple hypothesis testing). If the test is passed then the solution is returned. Otherwise the test must have encountered negative points within the box. We include these new negative points in $\sX^-$ and repeat  $\mathsf{FindMB}$ until the test is passed. To ensure that the levels for all our significance tests sum to no more than our desired level $\delta$, that is, $\sum_i \delta_i \leq \delta$, we have each successive test have significance level $\delta_i=\frac{\delta}{ i \log^2(i+1) \sum_{j=1}^\infty\frac{1}{j\log^2(j+1)}}$.  Note $\sum_{i=1}^\infty \delta_i=\delta$, and also that we must increment $i$ for each test (including every failed test during the merge process). Pseudocode is shown in Algorithm \ref{alg:solution}.

\textbf{Why is it necessary to solve the problem for subsets of dimensions?} The connection between $\mathsf{FindMB}$ and the maximum anchor box problem suggests a naive algorithm: sample a large number of points in the search space and find the maximum-box using $\mathsf{FindMB}$. Unfortunately, this approach fails in high dimensions: the size of the search space is simply too large to cover with samples, therefore, there are always large regions of space without sampled points that the anchor box can mistakenly include.  Thus, the resulting anchor box found by this naive procedure almost always has very low purity ($\rho \approx 0$). 

\textbf{Are the anchor boxes of a higher purity?}  Our method, by design, returns very high purity anchor boxes because it proposes regions without a single negative point in them and it applies a relatively weak statistical test (even a single negative point fails the test). As a result,  at $\rho=0.99$, the experimentally measured purity is usually at around $99.99\%$, which is much higher than required. 

\textbf{When is our algorithm optimal?}
An optimal algorithm for determining an anchor box would find the closest approximation to the maximum anchor box given that finding the true maximum is exponentially hard.  Our algorithm makes the implicit assumption that the anchor box of a set of features is included in the anchor box for a subset of these features. We call this the nestedness property. Denote the anchor box for features $\sI$ with $\sA_{\sI}$. $e$ is nested when $\forall$ features $\sI_1 \subseteq \sI_2. \sA_{\sI_2\text{max}}\subseteq \sA_{\sI_1\text{max}}$. The nestedness property holds for many $e$, for example when $D<3$ or when $e$ is linear, but it does not hold for all $e$. When the property does not hold, our algorithm still leads to a good solution but this solution may not be optimal.

If we had infinite samples, infinite computation and nestedness, we would find the maximal anchor box. However if we have finite samples and finite computation but the problem is nested, then we can find something close to optimal when we sample $N=\infty$ points to evenly cover the search space and the whole tree in $\mathsf{FindMB}$ is searched over $T=\infty$ iterations. The solution found in this case corresponds to $\rho=1$. If $e$ is not nested, we can find a good solution but cannot make any formal guarantees. In summary, our algorithm finds the largest anchor box in the limit $\rho\rightarrow 1$, $N\rightarrow \infty$ and $T\rightarrow \infty$, when $e$ is nested.

\begin{figure*}[t]
\centering
\begin{subfigure}{0.199\textwidth}
  \centering
  \includegraphics[width=\linewidth]{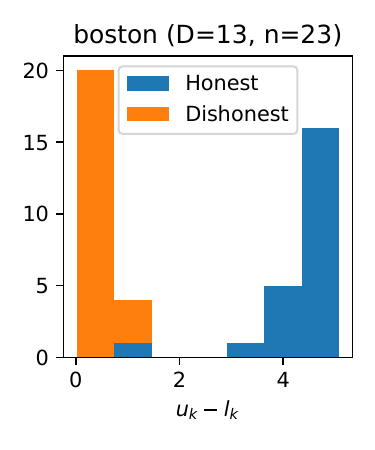}
\end{subfigure}%
\begin{subfigure}{0.199\textwidth}
  \centering
  \includegraphics[width=\linewidth]{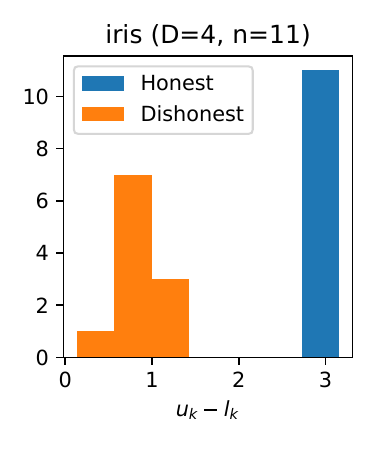}
\end{subfigure}%
\begin{subfigure}{0.199\textwidth}
  \centering
  \includegraphics[width=\linewidth]{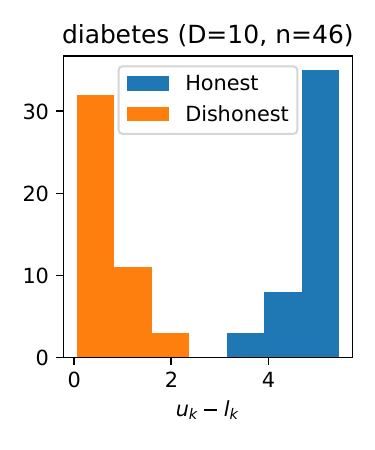}
\end{subfigure}%
\begin{subfigure}{0.199\textwidth}
  \centering
  \includegraphics[width=\linewidth]{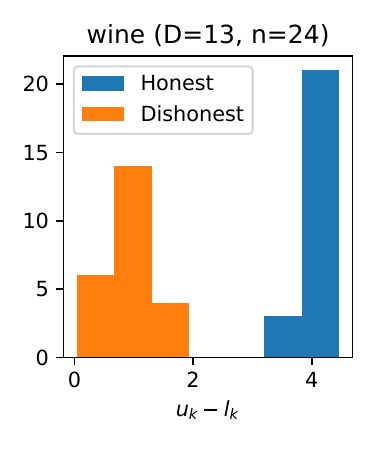}
\end{subfigure}%
\begin{subfigure}{0.199\textwidth}
  \centering
  \includegraphics[width=\linewidth]{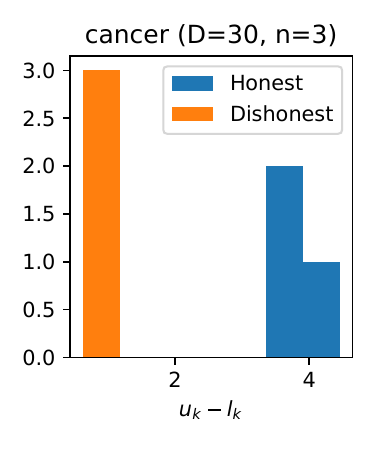}
\end{subfigure}
\vspace{-0.4cm}
\caption{We compute the anchor boxes for local linear explanations where feature $k$ is given 0 weight. In the case of the honest model, feature $k$ is not used for the prediction. In the case of the dishonest model, the prediction relies on feature $k$. We see a clear separation between the honest and dishonest models. The anchor box for the honest model is large along feature $k$, while for dishonest model, the size of the box is much smaller. } \label{fig:honesty}
\label{fig:test}
\end{figure*}

\textbf{Runtime complexity.} The complexity of $\mathsf{FindAnchor}$ is straightforward to analyse. It generates a tree of $\log_2(D)$ depth, and at each depth, it makes $D$ calls to $\mathsf{SolveRestricted}$ for a combined $O(D\log_2(D))$ calls. $\mathsf{SolveRestricted}$ is, however, more difficult to analyse. Due to the stochastic nature of our statistical test, a long runtime is possible, but unlikely. We expect it to rarely happen, because when the current best solution encounters a negative point it reduces the current best solution to the region excluding this point. This new region likely has high purity since it did not encounter a single negative point in the previous statistical test. In our experiments, $\mathsf{SolveRestricted}$ made on the order of 10-100 calls for $\mathsf{FindMB}$ noting that higher $|\sI|$ required more calls. The subroutine $\mathsf{FindMB}$ itself with $N$ positive points and $N$ negative points restricted to $T$ iterations has an $O(DNT)$ time complexity (in practice, the number of negative points may be higher than $N$, but we found in our experiments that it is usually on the order of magnitude $N$).

\section{Results and Discussion}
We present empirical results to support the efficacy of our proposed algorithm. We first showcase a comparison between the anchor boxes of honest versus dishonest local surrogate models. Next, we present quantitative results demonstrating that our algorithm captures a larger anchor box than baseline methods and that this larger volume covers more of the local cluster around the anchor point. 
Note that a larger volume is desirable since it means an explanation Finally, we explore the impact of different hyperparameter choices on the resulting anchor box and its computational cost. \footnote{Code is available at \url{https://github.com/dtak/anchor-box}.}
Our experiments use 5 tabular datasets available in the \texttt{scikit-learn} toolkit: boston (D=13, n=506), iris (D=4, n=150), diabetes (D=10, n=442), wine (D=13, n=178) and cancer (D=30, n=569). The model to be interpreted $f$ is a random forest classifier with the default hyperparameters (\texttt{n\_estimators=100}) (our method is model agnostic, results with neural networks are shown in Appendix \ref{app:nn}). Boston and Diabetes are regression datasets, so their task is to predict whether the outcome is above or below the median value. All features are standardised before training $f$. Our local surrogate models $g$ are either logistic regression or decision trees (\texttt{max\_depth=3}) as these models are commonly used for interpretability. We train them by sampling points from a Gaussian distribution centered on $a$ with covariance $\sigma^2I$ \citep{ribeiro2016should}. $\sigma^2$ is chosen to be maximal while requiring $g$ to be faithful for at least 99\% of the samples.

\subsection{Identifying when explanations are dishonest}
The motivating use case of our method is that given a surrogate model $g$, we want to know if $g$ can be trusted to extrapolate along each feature. We have a local linear explanation $g$ that claims that feature $k$ has 0 weight i.e. it does not contribute to the prediction. We then compute the anchor boxes against two models: an honest model $f_h$ that does not use feature $k$ and a dishonest model $f_d$ that does. We examine the size of the anchor boxes along feature $k$ to see if it possible to distinguish the two cases.

\begin{table*}[t]
\centering
\begin{adjustbox}{max width=\textwidth}
\begin{tabular}{llr@{$\pm$}lr@{$\pm$}lr@{$\pm$}lr@{$\pm$}lr@{$\pm$}lr@{$\pm$}lr@{$\pm$}lr@{$\pm$}lr@{$\pm$}lr@{$\pm$}l}
\toprule
 & & \multicolumn{4}{c}{boston (D=13)} & \multicolumn{4}{c}{iris (D=4)} & \multicolumn{4}{c}{diabetes (D=10)} & \multicolumn{4}{c}{wine (D=13)} & \multicolumn{4}{c}{cancer (D=30)} \\ 
     \cmidrule(r){3-6} \cmidrule(r){7-10} \cmidrule(r){11-14} \cmidrule(r){15-18} \cmidrule(r){19-22}
    &
    & \multicolumn{2}{c}{Vol ($\log_{10}$) $\uparrow$} & \multicolumn{2}{c}{Evals $\downarrow$} 
    & \multicolumn{2}{c}{Vol ($\log_{10}$) $\uparrow$} & \multicolumn{2}{c}{Evals $\downarrow$} 
    & \multicolumn{2}{c}{Vol ($\log_{10}$) $\uparrow$} & \multicolumn{2}{c}{Evals $\downarrow$} 
    & \multicolumn{2}{c}{Vol ($\log_{10}$) $\uparrow$} & \multicolumn{2}{c}{Evals $\downarrow$} 
    & \multicolumn{2}{c}{Vol ($\log_{10}$) $\uparrow$} & \multicolumn{2}{c}{Evals $\downarrow$} \\
\midrule
\multirow{3}{*}{\makecell[l]{Linear \\ surrogate}} & Radial & -5.7 & 8.6 & 55k & 9k & -0.7 & 1.2 & 56k & 4k & -2.7 & 5.3 & 57k & 8k & 1.8 & 2.7 & 64k & 3k & 4.6 & 14.2 & 67k & 7k \\                  
& Greedy Anchor & -0.4 & 5.7 & 678k & 37k & 0.8 & 0.7 & 187k & 6k & 0.4 & 3.1 & 498k & 36k & 3.7 & 2.4 & 688k & 0k & 6.2 & 13.6 & 1700k & 31k \\  
& Anchor (Ours) & \textbf{5.5} & 2.6 & 75k & 42k &\textbf{ 1.4} & 0.2 & 16k & 3k & \textbf{4.7} & 1.9 & 52k & 20k & \textbf{7.2} & 0.8 & 70k & 13k & \textbf{20.9} & 3.1 & 245k & 108k \\ 
\midrule
\multirow{3}{*}{\makecell[l]{Decision tree \\ surrogate}} & Radial & -4.2 & 7.7 & 57k & 9k & -0.3 & 1.5 & 58k & 5k & -1.3 & 4.0 & 59k & 6k & 2.7 & 2.7 & 65k & 3k & 4.6 & 13.5 & 67k & 6k \\             
& Greedy Anchor & 0.6 & 4.5 & 683k & 10k & 1.2 & 0.5 & 188k & 1k & 0.3 & 3.3 & 497k & 47k & 4.3 & 2.0 & 682k & 19k & 8.6 & 12.7 & 1686k & 67k \\    
& Anchor (Ours) & \textbf{6.9} & 0.8 & 51k & 10k & \textbf{1.7} & 0.3 & 13k & 4k & \textbf{5.6} & 0.6 & 38k & 8k & \textbf{7.5} & 0.7 & 59k & 14k & \textbf{21.5} & 3.0 & 206k & 90k \\
\bottomrule
\end{tabular}
\end{adjustbox}
\vspace{-0.3cm}
\caption{The volume of the guarantee region ($\rho=0.99$, $\delta=0.01$) captured by each method and the number of function evaluations $b$ they require. The means and standard deviations are shown over 20 test anchor points. Our method consistently results in a larger guarantee region than the baselines.} \label{tab:size}
\vspace{-0.3cm}
\end{table*}
\textbf{Experimental details.} We selected a target feature $k$ for each dataset that has the largest weight in a logistic regression model trained on the whole dataset. We then took a linear surrogate model $g$ that claimed not to use feature $k$ (this was achieved by masking feature $k$ with $a_x$ in $g$) and compute its anchor box against i) the honest model $f_h$ that did not use feature $k$ for prediction (we masked feature $k$ with $a_k$ again) and ii) the dishonest model $f_d$ that does use feature $k$ ($f_d$ is simply $f$). To compute the anchor box, we use $N=100$ positive points and at most $T=100$ iterations in $\mathsf{FindMB}$. The anchor points to evaluate were chosen from a test set of size 100. We filter any points distant from the decision boundary by requiring the confidence of $f$ to be at most 80\% and ensuring that $g$ is not faithful to $f_d$ for more than 30\% of the size of the bounding box at $a$ along feature $k$. For the remaining anchor points, $g$ is faithful to $f_h$ along feature $k$ but not faithful to $f_d$
\vspace{-0.3cm}
\begin{figure*}[t]
\begin{minipage}[H]{.6\textwidth}
\resizebox{\textwidth}{!}{
\begin{tabular}{llr@{$\pm$}lr@{$\pm$}lr@{$\pm$}lr@{$\pm$}l}
\toprule
  & & \multicolumn{4}{c}{Synthetic (D=2)} & \multicolumn{4}{c}{Synthetic (D=10)}\\ 
 
    \cmidrule(r){3-6} \cmidrule(r){7-10} 
    & & \multicolumn{2}{c}{\makecell[c]{\% of local \\ cluster $\uparrow$}} & \multicolumn{2}{c}{Evals $\downarrow$} 
    & \multicolumn{2}{c}{\makecell[c]{\% of local \\ cluster $\uparrow$}} & \multicolumn{2}{c}{Evals $\downarrow$}  \\
\midrule
\multirow{3}{*}{\makecell[l]{Linear \\ surrogate}} & Radial & 8.1 & 14.1 & 56k & 6k & 0.5 & 1.2 & 63k & 2k \\   
& Greedy Anchor & 66.5 & 20.9 & 85k & 5k & 13.6 & 12.9 & 516k & 0k \\    
& Anchor (Ours) & \textbf{82.0} & 13.6 & 7k & 3k & \textbf{53.1} & 19.5 & 54k & 9k \\ 
\midrule
\multirow{3}{*}{\makecell[l]{Decision tree \\ surrogate}} & Radial & 10.4 & 18.9 & 56k & 6k & 0.5 & 1.2 & 64k & 2k \\     
& Greedy Anchor & 58.0 & 27.3 & 82k & 8k & 12.3 & 12.5 & 508k & 20k \\                        
& Anchor (Ours) & \textbf{79.5} & 19.2 & 6k & 1k & \textbf{47.7} & 24.6 & 46k & 8k \\ 
\bottomrule
\end{tabular}}
\captionof{table}{The portion of the local cluster of points captured in the guarantee region ($\rho=0.99$, $\delta=0.01$) and the number of function evaluations $b$ they require. Synthetic dataset. The means and standard deviations are shown over 20 test anchor points. Our method is better at capturing the local cluster than the baselines.} \label{tab:synthetic}
\end{minipage} \hfill
\begin{minipage}[H]{0.31\textwidth}
\centering
\includegraphics[width=\textwidth]{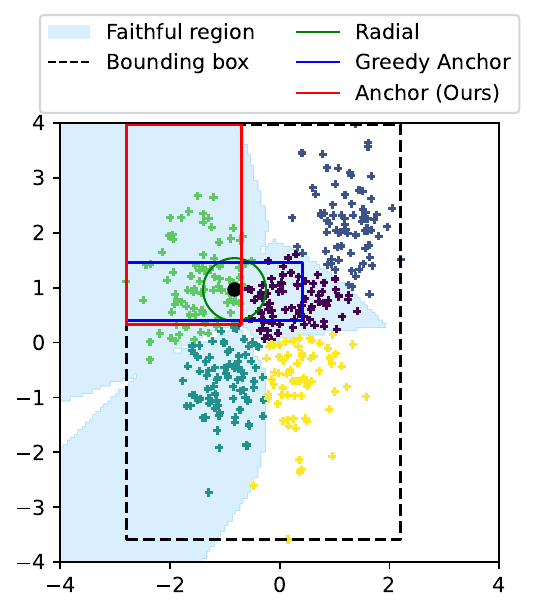} 
\vspace{-0.3cm}
\caption{Visualization of the Synthetic dataset at $D=2$. The anchor point is denoted by a black dot and it belongs to the green cluster of points. Our anchor box captures a large portion of the local cluster.}
\label{fig:synth2d}
\end{minipage} 
\end{figure*}


\textbf{Our anchor box clearly separates honest and dishonest explanations.} In Figure \ref{fig:honesty}, we see that the size of the anchor box is significantly smaller in the dishonest case than the honest ones on all datasets. The surrogate model extrapolates very little for the dishonest model along feature $k$ while it extrapolates far for honest model. Interestingly on the Boston dataset, the anchor box is quite small for one test point in the honest case. This can happen since our method does not always find the maximum anchor box, but importantly that the opposite is extremely unlikely. Our statistical guarantee ensures that the surrogate model is faithful within the resulting anchor box.
\vspace{-0.5cm}
\subsection{Capturing a larger volume with anchor boxes}
We show our anchor boxes capture a larger area than the baseline methods. Unfortunately, the most closely-related baselines work solely on categorical data. Hence, the first baseline we compare against is the radial explanation that captures points with distance at most $r$ around the anchor point. This is a natural way to find the guarantee region since the surrogate model is trained using symmetric Gaussian perturbations. The second baseline is a greedy approach for constructing an anchor box that starts with a small box around $a$ and expands its sides step-by-step while the guarantee is met.
\begin{table*}[t]
\centering
\begin{adjustbox}{max width=\textwidth}
\begin{tabular}{llr@{$\pm$}lr@{$\pm$}lr@{$\pm$}lr@{$\pm$}lr@{$\pm$}lr@{$\pm$}lr@{$\pm$}lr@{$\pm$}lr@{$\pm$}lr@{$\pm$}lr@{$\pm$}lr@{$\pm$}l}
\toprule
 & & \multicolumn{6}{c}{$T=$10} & \multicolumn{6}{c}{$T=$100} & \multicolumn{6}{c}{$T=$1000}\\ 
 
    \cmidrule(r){3-8} \cmidrule(r){9-14} \cmidrule(r){15-20} 
  &  & \multicolumn{2}{c}{Vol ($\log_{10}$)$\uparrow$} & \multicolumn{2}{c}{Evals $\downarrow$} & \multicolumn{2}{c}{Time (sec) $\downarrow$} 
    & \multicolumn{2}{c}{Vol ($\log_{10}$)$\uparrow$} & \multicolumn{2}{c}{Evals $\downarrow$} & \multicolumn{2}{c}{Time (sec) $\downarrow$} 
    & \multicolumn{2}{c}{Vol ($\log_{10}$)$\uparrow$} & \multicolumn{2}{c}{Evals $\downarrow$} & \multicolumn{2}{c}{Time (sec) $\downarrow$} \\
\midrule
\multirow{7}{*}{$D=10$} & $N=$10 & 4.9 & 0.3 & 116k & 29k & 2.1 & 0.7 & 5.0 & 0.2 & 113k & 23k & 2.1 & 0.5 & 5.0 & 0.1 & 102k & 16k & 1.9 & 0.3 \\ 
& $N=$30 & 5.2 & 0.1 & 99k & 15k & 2.4 & 0.6 & 5.1 & 0.1 & 99k & 18k & 3.4 & 2.1 & 5.2 & 0.1 & 99k & 20k & 3.7 & 3.2 \\ 
& $N=$100 & 5.2 & 0.1 & 99k & 12k & 3.2 & 0.5 & 5.2 & 0.0 & 116k & 27k & 16.2 & 11.4 & 5.3 & 0.1 & 110k & 21k & 58.6 & 97.0 \\ 
& $N=$300 & 5.3 & 0.0 & 106k & 18k & 4.5 & 0.8 & 5.3 & 0.0 & 136k & 19k & 30.2 & 12.1 & \multicolumn{6}{c}{Timeout} \\ 
&$N=$1000 & 5.3 & 0.0 & 134k & 16k & 6.3 & 1.1 & 5.3 & 0.0 & 165k & 16k & 43.9 & 11.5 & \multicolumn{6}{c}{Timeout} \\ 
& $N=$3000& 5.3 & 0.1 & 210k & 17k & 9.4 & 1.7 & 5.3 & 0.0 & 232k & 22k & 60.9 & 21.2 & \multicolumn{6}{c}{Timeout} \\
\cmidrule(r){2-20} 
& \multicolumn{19}{c}{$\rho=1$ optimal Vol ($\log_{10}$): 5.0 $\quad$ $\rho=0.99$ optimal Vol ($\log_{10}$): 6.2 }  \\
\midrule
\multirow{7}{*}{$D=30$} &$N=$10 & 22.3 & 0.5 & 598k & 56k & 14.8 & 1.6 & 22.2 & 0.4 & 592k & 36k & 14.8 & 1.0 & 22.0 & 0.7 & 628k & 78k & 16.0 & 2.4 \\ 
& $N=$30 & 23.0 & 0.3 & 632k & 54k & 20.6 & 2.4 & 23.0 & 0.3 & 642k & 73k & 28.7 & 8.4 & 23.0 & 0.3 & 634k & 71k & 32.9 & 14.9 \\ 
& $N=$100 & 23.5 & 0.1 & 644k & 57k & 31.6 & 3.6 & 23.6 & 0.1 & 749k & 71k & 149.3 & 24.0 & \multicolumn{6}{c}{Timeout} \\ 
& $N=$300 & 23.6 & 0.2 & 646k & 34k & 42.5 & 4.1 & 23.8 & 0.2 & 816k & 65k & 306.3 & 42.9 & \multicolumn{6}{c}{Timeout} \\ 
& $N=$1000 & 23.4 & 0.3 & 764k & 41k & 59.8 & 8.9 & 23.9 & 0.1 & 1013k & 71k & 617.0 & 105.6 & \multicolumn{6}{c}{Timeout} \\ 
& $N=$3000 & 23.4 & 0.3 & 1063k & 35k & 93.0 & 8.7 & \multicolumn{6}{c}{Timeout} & \multicolumn{6}{c}{Timeout} \\  
\cmidrule(r){2-20} 
& \multicolumn{19}{c}{$\rho=1$ optimal Vol ($\log_{10}$): 22.2 $\quad$ $\rho=0.99$ optimal Vol ($\log_{10}$): 33.6}  \\
\bottomrule
\end{tabular}
\end{adjustbox}
\vspace{-0.3cm}
\caption{Hyperparameter search at $\rho=0.99$ and $\delta=0.01$. $N$ is the number of positive points used in $\mathsf{FindMB}$ and $T$ is the number of search iterations. Note that the method is not expected to reach the optimal volume at $\rho=0.99$: the resulting anchor box has purity close to 99.99\% when it passes the statistical test. The means and standard deviations are shown over 20 test anchor points. The runtime is measured on a Tesla V100. } \label{tab:hyper_10}
\end{table*}

\textbf{Experimental details.} The radius $r$ of the radial explanation is determined by testing iteratively increasing values (up to 100 steps with logarithmic step-size) with our statistical test ($\rho=0.99$, $\delta=0.01$). The greedy anchor approach starts with a small box around the anchor point and increases its size along each axis stepwise as long as the statistical guarantee is met (up-to 100 steps in each direction with logarithmic step-size). We refer to this approach as greedy, since it is able to rapidly increase the size of the box, but it cannot escape a local-optima. Our approach uses $N=100$ positive points and at most $T=100$ iterations in $\mathsf{FindMB}$. The surrogate model $g$ is considered faithful if it predicts the same class as the model $f$, or if the difference in confidence between the class predicted by $f$ and $g$ is less than $10\%$.

\textbf{Our anchor boxes cover a significantly larger area than baseline methods.} Table \ref{tab:size} shows the results. We see that anchor boxes outperform the radial approaches in terms of the area covered and that our anchor box performs particularly well compared to the greedy approach capturing orders of magnitude larger area. Looking at the number of function evaluations $b$ needed, we see that our anchor box performs the best on 4 out of the 5 datasets with the radial approaches performing the best on cancer.
\vspace{-0.5cm}
\subsection{Capturing a local cluster using an anchor box}
While capturing a larger area is an indication that the explanation is more useful, we want to explicitly measure how much of the local cluster is captured by each explanation. Our goal is to show that our anchor box captures a larger portion of a local cluster than the baseline methods. For this, we use a synthetic dataset with Gaussian clusters of points so that we have access to the ground truth cluster allocations.

\textbf{Experimental details.} We used two variants: a simple 2-dimensional variant for illustration purposes and a more challenging, 10-dimensional variant. In each case, the dataset consists of 5 clusters of size 100. These clusters are generated from Gaussian distributions. The means of the generating Gaussians are drawn from $\mathcal{N}(0, 1)$ and their standard deviations are drawn from $U(0.3, 1)$ (with diagonal covariance matrices). $b$ is as defined in the previous experiment.

\textbf{Our anchor box captures a larger portion of the local cluster.} The results in Table \ref{tab:synthetic} show that our anchor box captures a higher portion of the local cluster than the baseline methods. For both $D=2$ and $D=10$, the radial method results in a very small region, greedy anchor finds a slightly larger region and our method yields the region with the best coverage. In Figure \ref{fig:synth2d}, we show an example run in the $D=2$ case. The anchor point is denoted with a black dot and it is part of the green cluster. The radial explanation covers a small area because it encounters an unfaithful region early on. Greedy anchor performs better, but it encounters a local optima that it cannot escape. Our method covers the most of the green cluster with the anchor box extending to all the way to the bounding box.
\vspace{-0.5cm}
\subsection{Hyperparameters}
Our method only has two hyperparameters: the number of positive points $N$ and the number of search iterations $T$ in $\mathsf{FindMB}$. We show how well the method performs with different hyperparameter settings on a synthetic dataset where the size of the optimal anchor box can be computed mathematically. We chose $b$ such that many of the features interact with each other, while many features also do not contribute to the outcome:
\begin{equation}
b(x)=
\begin{cases}
1 \quad \text{if} \, \sum_{i=1}^\frac{D}{2} |x_i| < \frac{D}{4}  \\
0 \quad \text{otherwise} \,.
\end{cases}
\end{equation}
Table \ref{tab:hyper_10} summarizes the results. We see that the compute time increases with both $N$ and $T$ with good trade-off being offered in the $N=30-300$ and $T=10-100$ region.


\section{Conclusions}
We presented a method for finding the guarantee region for a local surrogate model. In the resulting axis-aligned box, the local surrogate model is guaranteed to be faithful to the complex model, which is confirmed by statistical testing. Our method tackles a significant computational challenge by employing a divide-and-conquer strategy over the input features. Empirical results confirm that our method is able to identify misleading explanations by presenting a significantly reduced guarantee region. Moreover, our method outperforms the baselines both in the size of the guarantee region and in the coverage of the local data cluster.




\bibliography{aistats2024.bib}
\appendix
\section{Theoretical results}

\textbf{Theorem 3.2}
If there exists a true function $e_0$ for the data and a set of possible faithfulness functions $\mathcal{E}=\{e_k: \R^D\rightarrow \{0, 1\}\}_{k=1}^K$ with $K={D \choose \lfloor \frac{D}{2} \rfloor}$ that agree on a collection of input points, but have different bounding boxes and  $\frac{|\sA_\text{max}(x^*, e_0, \rho=1)|}{|\sA_\text{max}(x^*, e_k, \rho=1)|} = r$, where $0<r<1$ is the ratio of volumes of maximal bounding boxes for different $e_k$, then for $L$ evaluations, $x^{(1)} \dots x^{(L)}$, 
\begin{align}
    |\{e_k \in \mathcal{E}|\forall x^{(l)}. e_0(x^{(l)}) = e_k(x^{(l)}) \}| \geq K-L
\end{align} 

\begin{proof}
\label{app:proof}

Let $\sS_1 \dots \sS_{{D \choose \lfloor \frac{D}{2} \rfloor}}$ be the $\lfloor \frac{D}{2} \rfloor$ size subsets of $\{1 \dots D\}$.
Let 
\begin{equation}
e_0(x)=
\begin{cases}
1 \quad \text{if} \, |\{x_d \geq r\}| \leq \lfloor \frac{D}{2} \rfloor - 1 \wedge \forall d. 0 \leq x_d \leq 1\\
0 \quad \text{otherwise} \,,
\end{cases}
\end{equation}
and for $1 \leq k \leq K$,
\begin{equation}
e_k(x)=
\begin{cases}
1 \quad \text{if} \, d\in \sS_k. 0 \leq x_d \leq 1 \wedge d\notin \sS_k. 0 \leq x_d \leq r \\
1 \quad \text{if} \, e_0(x)=1\\
0 \quad \text{otherwise} \,.
\end{cases}
\end{equation}
We immediately see that $e_0(x)=1 \implies \forall k. e_k(x)=1$. Let $1 \leq i \leq K$ and have $e_i(x) \neq e_0(x)$. Using $e_0(x)=1 \implies \forall k. e_k(x)=1$, we have $e_i(x)=1 \wedge e_0(x)=0$. This is only possible when exactly ${D \choose \lfloor \frac{D}{2} \rfloor}$ dimensions of $x$
are between $r$ and $1$, precisely the dimensions in $\sS_i$. Since for $1 \leq j\neq i \leq K$, $\sS_i\neq \sS_j$, we have $e_0(x)=e_j(x)$. As a result, for  $1 \leq j\neq i \leq K. e_0(x)\neq e_i(x) \implies e_0(x)=e_j(x)$.

Given a sequence $x^{(1)} \dots x^{(L)}$, each $x^{(l)}$ may only show disagreement with $e_0$ with only one $e_k$, therefore the number of $e_k$ that agree with $e_0$ on all $x^{(1)} \dots x^{(L)}$ is at least $K-L$:
\begin{equation}
    |\{e_k \in \mathcal{E}|\forall x^{(l)}. e_0(x^{(l)}) = e_k(x^{(l)}) \}| \geq K-L
\end{equation}  

Finally, we need to show that $\frac{|\sA_\text{max}(a, e_0)|}{|\sA_\text{max}(a, e_k)|} = r$. 

For $e_0$, the maximum anchor box has $l_d=0$ and $u_d=r$ for  $1 \leq d \leq \frac{D}{2} + 1$ and $u_d=1$ for $\frac{D}{2} + 1 < d \leq D$ with size $|\sA_\text{max}(x^*, e_0, \rho=1)|=r^{\frac{D}{2} + 1}$ (the maximum box is not unique, any $u$ with $\frac{D}{2} + 1$ entries containing $r$ and $\frac{D}{2} - 1$ entries containing $1$ achieves the maximum size).

For $e_k$, the maximum box is unique at $l_d=0$ and $u_d=r$ for  $d \notin \sS_n$ and $u_d=1$ for $d\in \sS_k$ with size $|\sA_\text{max}(x^*, e_k, \rho=1)|=r^\frac{d}{2}$.

Therefore $\frac{|\sA_\text{max}(x^*, e_0)|}{|\sA_\text{max}(x^*, e_k)|} = r$ concluding the proof.
\end{proof}

\section{Box expansion} \label{app:expansion}

When $\mathsf{FindMB}$ finds the maximum box covering the positive points, we need to expand its sides so that it is spanned by negative points. An algorithmic choice we made is to expand the sides that yield the largest size improvement first. Here, we show that this choice has a minor effect on the size of the resulting anchor box. Tables \ref{tab:border_1} and \ref{tab:border_2} show the comparison against the random order. 

\begin{table*}[t]
\centering
\begin{adjustbox}{max width=\textwidth}
\begin{tabular}{llr@{$\pm$}lr@{$\pm$}lr@{$\pm$}lr@{$\pm$}lr@{$\pm$}lr@{$\pm$}lr@{$\pm$}lr@{$\pm$}lr@{$\pm$}lr@{$\pm$}l}
\toprule
 & & \multicolumn{4}{c}{boston (D=13)} & \multicolumn{4}{c}{iris (D=4)} & \multicolumn{4}{c}{diabetes (D=10)} & \multicolumn{4}{c}{wine (D=13)} & \multicolumn{4}{c}{cancer (D=30)} \\ 
     \cmidrule(r){3-6} \cmidrule(r){7-10} \cmidrule(r){11-14} \cmidrule(r){15-18} \cmidrule(r){19-22}
    &
    & \multicolumn{2}{c}{Vol ($\log_{10}$) $\uparrow$} & \multicolumn{2}{c}{Evals $\downarrow$} 
    & \multicolumn{2}{c}{Vol ($\log_{10}$) $\uparrow$} & \multicolumn{2}{c}{Evals $\downarrow$} 
    & \multicolumn{2}{c}{Vol ($\log_{10}$) $\uparrow$} & \multicolumn{2}{c}{Evals $\downarrow$} 
    & \multicolumn{2}{c}{Vol ($\log_{10}$) $\uparrow$} & \multicolumn{2}{c}{Evals $\downarrow$} 
    & \multicolumn{2}{c}{Vol ($\log_{10}$) $\uparrow$} & \multicolumn{2}{c}{Evals $\downarrow$} \\
\midrule
\multirow{2}{*}{Linear surrogate} 
& Anchor (Largest expansion first) & 5.5 & 2.6 & 75k & 42k & 1.4 & 0.2 & 16k & 3k & 4.7 & 1.9 & 52k & 20k & 7.2 & 0.8 & 70k & 13k & 20.9 & 3.1 & 245k & 108k \\ 
& Anchor (Random order) & 5.5 & 2.3 & 65k & 28k & 1.3 & 0.3 & 14k & 3k & 4.6 & 1.8 & 46k & 17k & 6.9 & 0.6 & 67k & 11k & 19.6 & 3.6 & 218k & 93k \\
\midrule
\multirow{2}{*}{Decision tree surrogate} 
& Anchor (Largest expansion first) & 6.9 & 0.8 & 51k & 10k & 1.7 & 0.3 & 13k & 4k & 5.6 & 0.6 & 38k & 8k & 7.5 & 0.7 & 59k & 14k & 21.5 & 3.0 & 206k & 90k \\
& Anchor (Random order) & 6.7 & 0.7 & 46k & 9k & 1.7 & 0.3 & 11k & 2k & 5.3 & 0.8 & 38k & 10k & 7.3 & 0.7 & 56k & 15k & 20.1 & 3.9 & 177k & 66k \\ 
\bottomrule
\end{tabular}
\end{adjustbox}
\caption{The volume of the guarantee region ($\rho=0.99$, $\delta=0.01$) captured by expansion method. The means and standard deviations are shown over 20 test anchor points. The largest expansion first order yields a minor improvement over random order.} \label{tab:border_1}
\end{table*}

\begin{table*}[t]
\centering
\begin{adjustbox}{max width=\textwidth}
\begin{tabular}{llr@{$\pm$}lr@{$\pm$}lr@{$\pm$}lr@{$\pm$}l}
\toprule
  & & \multicolumn{4}{c}{Synthetic (D=2)} & \multicolumn{4}{c}{Synthetic (D=10)}\\ 
 
    \cmidrule(r){3-6} \cmidrule(r){7-10} 
    & & \multicolumn{2}{c}{\makecell[c]{\% of local \\ cluster $\uparrow$}} & \multicolumn{2}{c}{Evals $\downarrow$} 
    & \multicolumn{2}{c}{\makecell[c]{\% of local \\ cluster $\uparrow$}} & \multicolumn{2}{c}{Evals $\downarrow$}  \\
\midrule
\multirow{2}{*}{\makecell[l]{Linear \\ surrogate}}
& Anchor (Largest expansion first) & 82.0 & 13.6 & 7k & 3k & 53.1 & 19.5 & 54k & 9k \\ 
& Anchor (Random order) & 81.0 & 14.0 & 7k & 1k & 52.4 & 20.3 & 51k & 8k \\    
\midrule
\multirow{2}{*}{\makecell[l]{Decision tree \\ surrogate}}                        
& Anchor (Largest expansion first) & 79.5 & 19.2 & 6k & 1k & 47.7 & 24.6 & 46k & 8k \\ 
& Anchor (Random order) & 79.6 & 16.1 & 6k & 1k & 47.6 & 21.9 & 44k & 9k \\
\bottomrule
\end{tabular}
\end{adjustbox}
\caption{The coverage of the guarantee region ($\rho=0.99$, $\delta=0.01$) for each expansion order. The means and standard deviations are shown over 20 test anchor points. The largest expansion first order is on-par with the random order.} \label{tab:border_2}
\end{table*}

\section{Results on neural networks} \label{app:nn}

The main paper showcases experiments with random forest classifiers. Here, we show that the results are very similar when the back-box model is a neural network. Tables \ref{tab:nn_vol} and \ref{tab:nn_cap} replicate the experiment from Tables \ref{tab:size} and \ref{tab:synthetic}, but with deep neural networks instead of random forests as the black-box model $f$ to be interpreted.

\begin{table*}[t]
\centering
\begin{adjustbox}{max width=\textwidth}
\begin{tabular}{llr@{$\pm$}lr@{$\pm$}lr@{$\pm$}lr@{$\pm$}lr@{$\pm$}lr@{$\pm$}lr@{$\pm$}lr@{$\pm$}lr@{$\pm$}lr@{$\pm$}l}
\toprule
 & & \multicolumn{4}{c}{boston (D=13)} & \multicolumn{4}{c}{iris (D=4)} & \multicolumn{4}{c}{diabetes (D=10)} & \multicolumn{4}{c}{wine (D=13)} & \multicolumn{4}{c}{cancer (D=30)} \\ 
     \cmidrule(r){3-6} \cmidrule(r){7-10} \cmidrule(r){11-14} \cmidrule(r){15-18} \cmidrule(r){19-22}
    &
    & \multicolumn{2}{c}{Vol ($\log_{10}$) $\uparrow$} & \multicolumn{2}{c}{Evals $\downarrow$} 
    & \multicolumn{2}{c}{Vol ($\log_{10}$) $\uparrow$} & \multicolumn{2}{c}{Evals $\downarrow$} 
    & \multicolumn{2}{c}{Vol ($\log_{10}$) $\uparrow$} & \multicolumn{2}{c}{Evals $\downarrow$} 
    & \multicolumn{2}{c}{Vol ($\log_{10}$) $\uparrow$} & \multicolumn{2}{c}{Evals $\downarrow$} 
    & \multicolumn{2}{c}{Vol ($\log_{10}$) $\uparrow$} & \multicolumn{2}{c}{Evals $\downarrow$} \\
\midrule
\multirow{3}{*}{Linear surrogate} & Radial & -1.4 & 6.6 & 60k & 7k & 1.5 & 0.7 & 65k & 2k & 1.7 & 2.6 & 64k & 3k & 4.4 & 2.8 & 67k & 3k & 7.9 & 14.7 & 68k & 7k \\
& Greedy Anchor & -0.9 & 5.3 & 665k & 68k & 1.2 & 0.3 & 188k & 2k & 2.3 & 1.4 & 516k & 0k & 3.3 & 2.0 & 683k & 18k & 4.0 & 17.1 & 1643k & 216k \\
& Anchor (Ours) & 5.7 & 1.1 & 73k & 21k & 1.6 & 0.3 & 17k & 5k & 4.9 & 1.2 & 55k & 13k & 6.6 & 0.7 & 95k & 20k & 18.8 & 4.0 & 269k & 129k \\
\midrule
\multirow{3}{*}{Decision tree surrogate} & Radial & -1.7 & 6.0 & 60k & 6k & -0.3 & 1.6 & 58k & 6k & -0.6 & 4.2 & 60k & 6k & 3.2 & 2.9 & 66k & 3k & 5.1 & 16.0 & 67k & 8k \\
& Greedy Anchor & -0.2 & 4.3 & 676k & 27k & 0.5 & 0.4 & 187k & 4k & 0.8 & 2.7 & 505k & 35k & 2.7 & 2.4 & 681k & 25k & 4.0 & 17.1 & 1633k & 212k \\
& Anchor (Ours) & 6.0 & 0.9 & 77k & 29k & 1.2 & 0.1 & 27k & 4k & 5.3 & 0.6 & 54k & 11k & 6.6 & 0.5 & 106k & 30k & 19.2 & 4.0 & 280k & 145k \\

\bottomrule
\end{tabular}
\end{adjustbox}
\caption{The volume of the guarantee region ($\rho=0.99$, $\delta=0.01$) captured by each method and the number of function evaluations $b$ they require. The base model is a neural network with one hidden layer of width 100 with ReLU activations. The means and standard deviations are shown over 20 test anchor points.} \label{tab:nn_vol}
\end{table*}

\begin{table*}[t]
\centering
\begin{adjustbox}{max width=\textwidth}
\begin{tabular}{llr@{$\pm$}lr@{$\pm$}lr@{$\pm$}lr@{$\pm$}l}
\toprule
  & & \multicolumn{4}{c}{Synthetic (D=2)} & \multicolumn{4}{c}{Synthetic (D=10)}\\ 
 
    \cmidrule(r){3-6} \cmidrule(r){7-10} 
    & & \multicolumn{2}{c}{\makecell[c]{\% of local \\ cluster $\uparrow$}} & \multicolumn{2}{c}{Evals $\downarrow$} 
    & \multicolumn{2}{c}{\makecell[c]{\% of local \\ cluster $\uparrow$}} & \multicolumn{2}{c}{Evals $\downarrow$}  \\
\midrule
\multirow{3}{*}{\makecell[l]{Linear \\ surrogate}} & Radial & 10.2 & 17.0 & 58k & 5k & 2.7 & 6.4 & 65k & 1k \\
& Greedy Anchor & 66.4 & 20.5 & 86k & 3k & 9.0 & 8.7 & 510k & 20k \\
& Anchor (Ours) & 79.0 & 21.4 & 7k & 2k & 31.7 & 17.0 & 88k & 19k \\

\midrule
\multirow{3}{*}{\makecell[l]{Decision tree \\ surrogate}} & Radial & 8.1 & 17.0 & 57k & 5k & 1.7 & 5.6 & 65k & 1k \\
& Greedy Anchor & 57.9 & 27.0 & 82k & 7k & 4.0 & 4.3 & 506k & 14k \\
& Anchor (Ours) & 79.0 & 16.8 & 7k & 2k & 30.1 & 16.6 & 92k & 17k \\
\bottomrule
\end{tabular}
\end{adjustbox}
\caption{The coverage of the guarantee region ($\rho=0.99$, $\delta=0.01$) captured by each method and the number of function evaluations $b$ they require. The base model is a neural network with one hidden layer of width 100 with ReLU activations. The means and standard deviations are shown over 20 test anchor points.} \label{tab:nn_cap}
\end{table*}

\end{document}